\documentclass{article}

\usepackage[final]{neurips_2019}

\usepackage[utf8]{inputenc} 
\usepackage[T1]{fontenc}    
\usepackage{hyperref}       
\usepackage{url}            
\usepackage{booktabs}       
\usepackage{amsfonts}       
\usepackage{nicefrac}       
\usepackage{microtype}      

\usepackage[final,inline,nomargin,index]{fixme}
\fxsetup{theme=color,mode=multiuser,inlineface=\itshape,envface=\itshape}
\FXRegisterAuthor{jw}{ajw}{\colorbox{gray!10!white}{\color{black}Jon}}
\FXRegisterAuthor{sf}{asf}{\colorbox{blue!10!white}{\color{black}Sorelle}}
\FXRegisterAuthor{ss}{ass}{\colorbox{red!10!white}{\color{black}Silvia}}
\FXRegisterAuthor{kl}{akl}{\colorbox{red!10!white}{\color{black}Kadan}}

\newcommand{\co}{CO$_2$}
\newcommand{\report}{Energy Usage Report}
\newcommand{\energyusage}{\texttt{energyusage}}

\usepackage{multicol}
\usepackage{tcolorbox}
\usepackage{amsmath}

\AtBeginDocument{%
  \providecommand\BibTeX{{%
    \normalfont B\kern-0.5em{\scshape i\kern-0.25em b}\kern-0.8em\TeX}}}

\title{Energy Usage Reports: Environmental awareness as part of algorithmic accountability\thanks{This work is funded in part by the Mozilla Responsible Computer Science Challenge and by the NSF under grant IIS-1633387.}}

\author{Kadan Lottick \thanks{Both authors contributed equally to this research.}\\
Haverford College\\
Haverford, Pa, USA\\
\texttt{klottick@haverford.edu}\\
\And
Silvia Susai $^\dagger$\\
Haverford College\\
Haverford, Pa, USA\\
\texttt{ssusai@haverford.edu}\\
\And
Sorelle A. Friedler\\
Haverford College\\
Computer Science Dept.\\
Haverford, Pa, USA\\
\texttt{sorelle@cs.haverford.edu}\\
\And
Jonathan P. Wilson\\
Haverford College\\
Environmental Studies Dept.\\
Haverford, Pa, USA\\
\texttt{jwilson@haverford.edu}\\
}

\begin{document}

\maketitle

\begin{abstract}
The carbon footprint of algorithms must be measured and transparently reported so computer scientists can take an honest and active role in environmental sustainability. In this paper, we take analyses usually applied at the industrial level and make them accessible for individual computer science researchers with an easy-to-use Python package. Localizing to the energy mixture of the electrical power grid, we make the conversion from energy usage to CO$_2$ emissions, in addition to contextualizing these results with more human-understandable benchmarks, such as automobile miles driven. We also include comparisons with energy mixtures employed in electrical grids around the world. We propose including these automatically-generated \emph{Energy Usage Reports} as part of standard algorithmic accountability practices, and demonstrate the use of these reports as part of model-choice in a machine learning context.
\end{abstract}

\section{Introduction}

Anthropogenic climate change is a global environmental problem caused by human-induced changes to the global carbon cycle through emissions of greenhouse gases, particularly carbon dioxide (CO$_2$) (\cite{IPCC_WG1_SFP, IPCC_WG1_anthropogenic_forcing}). Carbon dioxide emissions change the Earth's energy balance: a higher concentration of CO$_2$ in the atmosphere increases the amount of longwave radiation absorbed by the atmosphere and reradiated to the surface, increasing global average surface temperatures and altering long-term climatic trends (\cite{IPCC_WG1_TechnicalSummary, IPCC_WG1_anthropogenic_forcing}). In most industrialized countries, the energy sector is one of the top contributors to national CO$_2$ emissions (\cite{IPCC_WG1_anthropogenic_forcing}). 
Minimizing carbon emissions through the use of increasingly efficient computational methods is an appropriate response to climate change under the framework of Environmental Justice (\cite{IPCC_WG2_resilience}). Computer scientists could play a central role in reducing carbon emissions economy-wide through the design and implementation of more energy-efficient algorithms.

There have been many attempts to measure the carbon footprint of computing (\cite{strubell2019energy,arif_mahmood_cloud_computing,posani_envi_cloud_storage,ensmenger_longpiece,ensmenger_computation,coroama_energy_intensity_internet}), with special attention paid to energy-intensive operations such as Bitcoin mining and cloud computing (\cite{bitcoin_article,baliga_green_cloud_computing,arif_mahmood_cloud_computing,posani_envi_cloud_storage}). However, the details of electricity generation and power consumption make it difficult to translate from an individual device's operations to the energy use and carbon emissions which result from those operations. 

In this paper, we introduce a Python package \energyusage\footnote{\texttt{pip install energyusage}: \url{http://github.com/responsibleproblemsolving/energy-usage}} that can calculate the energy and \co\ emissions of a given function as well as output an \report\ giving context to these results. We argue that any attempts to use algorithms to tackle climate change should also report on the algorithms' direct emissions impacts. 

\section{\report}

\begin{figure*}
\begin{tcolorbox}
\centering

\textbf{\report}\\
Energy usage and \co\ emissions for the function \texttt{exp} with input \texttt{10}.

\vspace{-.1in}

\begin{multicols}{2}
\textbf{Energy Usage Readings}\\~\\
\begin{tabular}{rl}
Average baseline wattage: & 2.35 watts \\
Average total wattage: & 15.53 watts\\
Average process wattage: &  13.18 watts\\
Process duration: & 0:16:40\\

\end{tabular}
\columnbreak

\textbf{Energy Mix Data}\\
\includegraphics[width=1.5in]{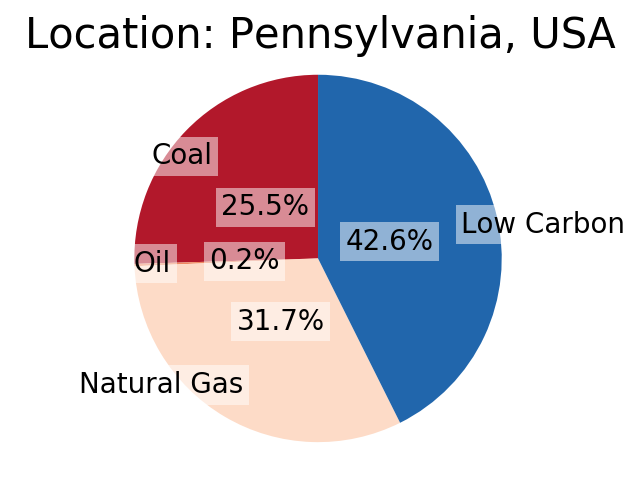}
\end{multicols}

\vspace{-.1in}
\begin{tcolorbox}[width=3.5in, colback=white]
\begin{center}
\vspace{-.05in}
\begin{tabular}{rl}
\textbf{Total kilowatt hours used:} & \large{0.00367 kWh}\\
\textbf{Effective emissions}: & \large{1.78e-03 kg \co}\\
\end{tabular}
\vspace{-.05in}
\end{center}
\end{tcolorbox}
\vspace{-.1in}

\begin{multicols}{2}
\textbf{Assumed Carbon Equivalencies}\\
\begin{tabular}{rl}
Coal:     &   996 kg \co/MWh\\
Oil: & 817 kg \co/MWh\\
Natural gas: &744 kg \co/MWh\\
Low carbon: & 0 kg \co/MWh
\end{tabular}
\columnbreak

\textbf{\co\ Emissions Equivalents}\\
\begin{tabular}{rl}
Miles driven: & 7.26 e-10 mi\\
Min. of 32-in. LCD TV: 
& 1.10 min.\\
\% of \co\ per 
US house
/day: &  5.84 e-10\% 
\end{tabular}
\end{multicols}

\vspace{-.2in}
\begin{center}
\textbf{Emission Comparisons}\\
\co\ emissions for the function if the computation had been performed elsewhere.
\end{center}
\includegraphics[width=1.5in]{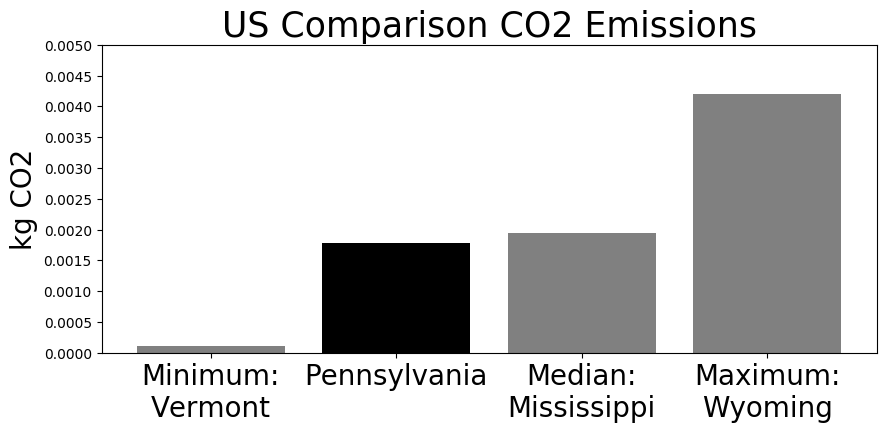}~~
\includegraphics[width=1.5in]{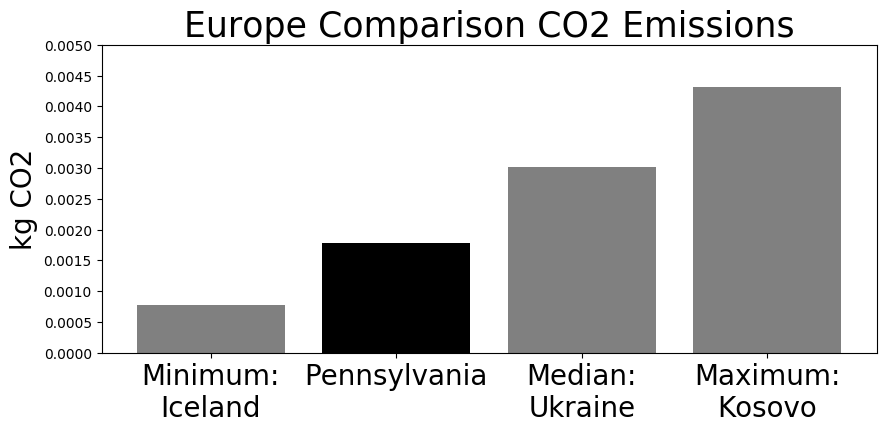}~~
\includegraphics[width=1.5in]{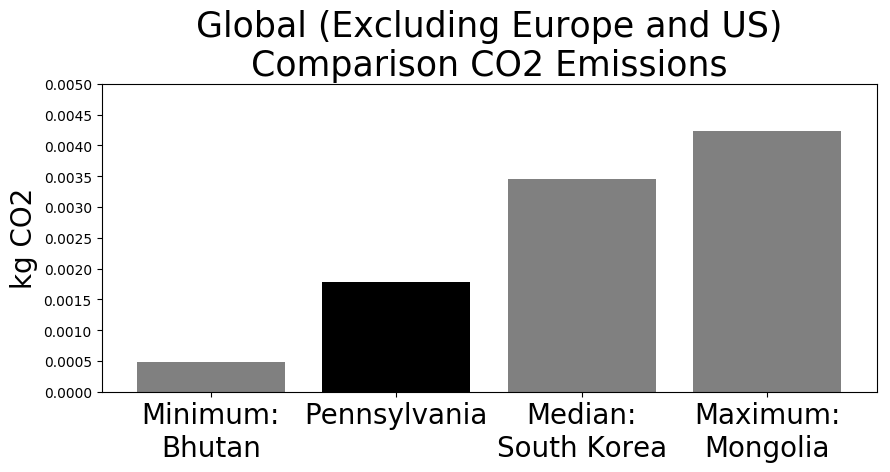}

\end{tcolorbox}

\caption{An example \report\ for a simple exponential function (code in Figure \ref{fig:code}).}
\label{fig:report}
\end{figure*}

When considering the environmental impact of algorithms, it is generally acknowledged that reporting the energy usage and \co\ emissions is the ``gold standard" (\cite{schwartz2019green, strubell2019energy}); it is the standard we adopt here. Efficiency and energy usage depends on an entire pipeline of energy generation, from individual computer power loss rates to the balance of the power grid between renewable energy sources, such as wind energy, versus fossil fuel sources, such as coal. Energy source mixtures are different by location, so reporting emissions statistics requires knowledge of where an algorithm was run as well as the specific energy mixture data for that location. We concretely address these real-world concerns, which we believe are at the heart of the goals of these questions. We accompany the resulting measurements with human-understandable benchmarks, such as equivalent miles driven, that make the environmental impact tangible.

An example report for a simple exponential function is given in Figure \ref{fig:report}. The report is designed to be self-explanatory, but a full explanation of the rationale behind each section is given in Appendix \ref{sec:report_desc}. The basic methodology uses the RAPL interface to directly measure a machine's energy usage and adjusts that based on assumed power loss to determine kilowatt hours used. A location API and accompanying per-US state and per-country energy mix data are then used, along with \co\ emissions rates for fuel sources, to determine the \co\ emissions of the algorithm.

Because of wide variance in published electricity grid carbon emission rates (even for the same fuel sources), we elected to reverse-engineer the formulas for the carbon footprint of each individual \textit{fuel source} from the state emission output and energy mix eGRID data, which have been found to be reliable.  We then applied that formula to the energy mix of each country (see Appendix \ref{sec:method_emissions} for further details). This approach not only provides us additional consistency between the datasets, but it also ensures that the conversions pertain, specifically, to electricity consumption.
The full methodology, validation, and package usage descriptions are given in the Appendix.

\section{Example Energy Usage Analysis: Machine Learning Algorithms}

One important algorithmic domain that motivates much of the recent interest in energy usage analysis (e.g., \cite{strubell2019energy, schwartz2019green}) is that of machine learning. In addition to previously surfaced concerns about the overall energy consumption of machine learning training algorithms (\cite{strubell2019energy}) we also argue here that energy usage is a distinct trade-off from those traditionally studied in machine learning (such as precision vs. recall or fairness vs. accuracy). While our initial hypothesis was that accuracy (the percent of correct predictions) 
would increase as energy usage 
increased, thus somewhat justifying the energy usage of more complex models, the reality we found was not this simple. The experimental setup is given in Appendix \ref{sec:experimental_setup}.

\begin{figure}
    \centering
    \includegraphics[width=2.7in]{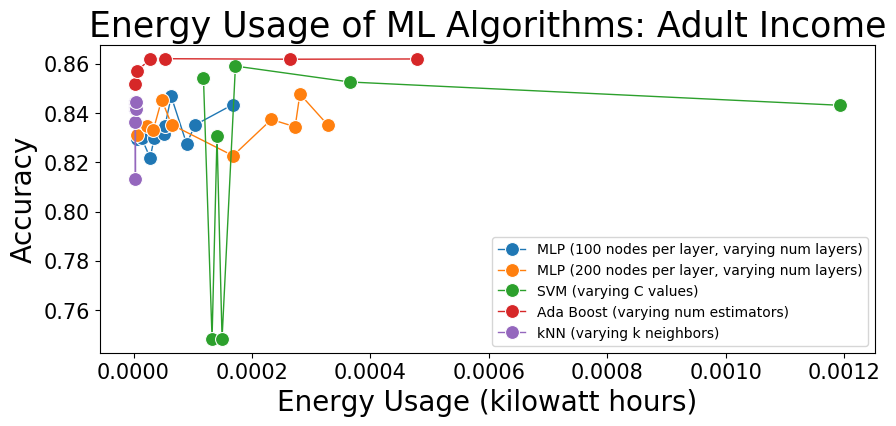}~
    \includegraphics[width=2.7in]{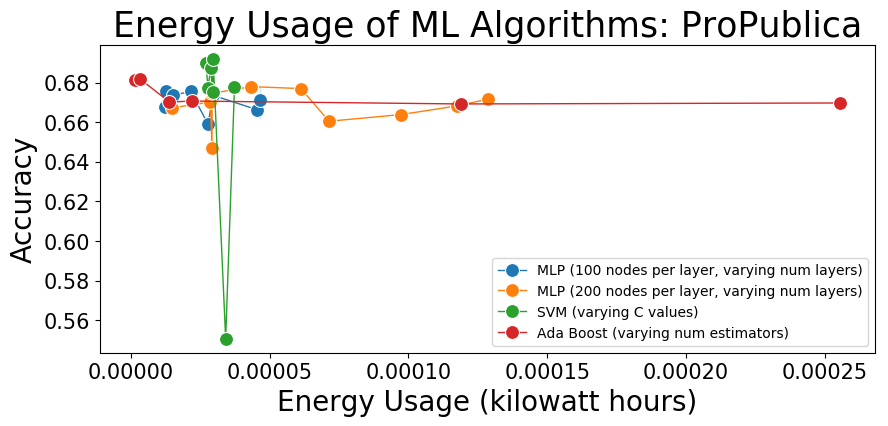}
    \caption{The energy usage and accuracy results when training four model types and varying based on a tunable parameter.  The multi-layer perceptron (MLP) classifier was considered with hidden layers from 1 to 10 at integral increments for architectures with both 100 nodes per layer and 200 nodes per layer.  The support vector machine (SVM) was trained with an RBF kernel and C values varying from $1e-3$ to $1000$ by powers of 10.  The $k$-nearest neighbor (kNN) classifier was trained for values of $k \in [5, 30]$ at increments of 5 as well as for $k=1$.  Ada Boost (with a decision tree base estimator) was trained for varying numbers of estimators: $[50, 100, 500, 1000, 5000, 10000]$.  Results are shown for both the Adult Income and ProPublica data sets.  The results demonstrate that energy usage is measuring a new, distinct trade-off.}
    \label{fig:ml}
\end{figure}

The results (see Figure \ref{fig:ml}) demonstrate that energy usage is a distinct measurement from those usually considered when choosing models, and should be analyzed and optimized for in addition to traditional measures. The multi-layer perceptron 
increases energy usage with additional layers, but the impact on accuracy is less consistent. Increasing the number of nodes in the hidden layers increases energy usage without necessarily increasing accuracy (and in some cases decreasing it in comparison to the multi-layer perceptron with fewer nodes per layer). Thus, in addition to the search for a good architecture taking substantial  
energy usage 
(\cite{strubell2019energy}), some architectures are themselves more or less energy intensive. 
The $k$-nearest neighbor classifier essentially doesn't vary in terms of energy usage for varying values of $k$, but changes substantially in accuracy.

Many of these results are not surprising when considering the details of the algorithms with respect to energy usage, but we argue that it is exactly this extra consideration that is warranted. Models that may be repeatedly retrained (e.g., as data is updated) should have the energy usage examined as one of the standard measures to optimize. Yet this also presents a potential contradiction and challenge, since in order to examine this tradeoff, further energy must be expended. Still, some algorithms, such as the $k$-nearest neighbor classifier, perform well with consistently lower energy usage.\footnote{In fact, the $k$-nearest neighbor classifier had energy usage that was too low to be measured using the RAPL interface on the ProPublica data.} We hope that transparency as to the energy usage of models will lead to further focus on and development of such low energy machine learning algorithms.

\bibliographystyle{ACM-Reference-Format}
\bibliography{energyusage}

\newpage

\appendix

\section{Background: Energy Measurement}
\label{sec:background}

\begin{figure}
    \centering
    \includegraphics[width=3in]{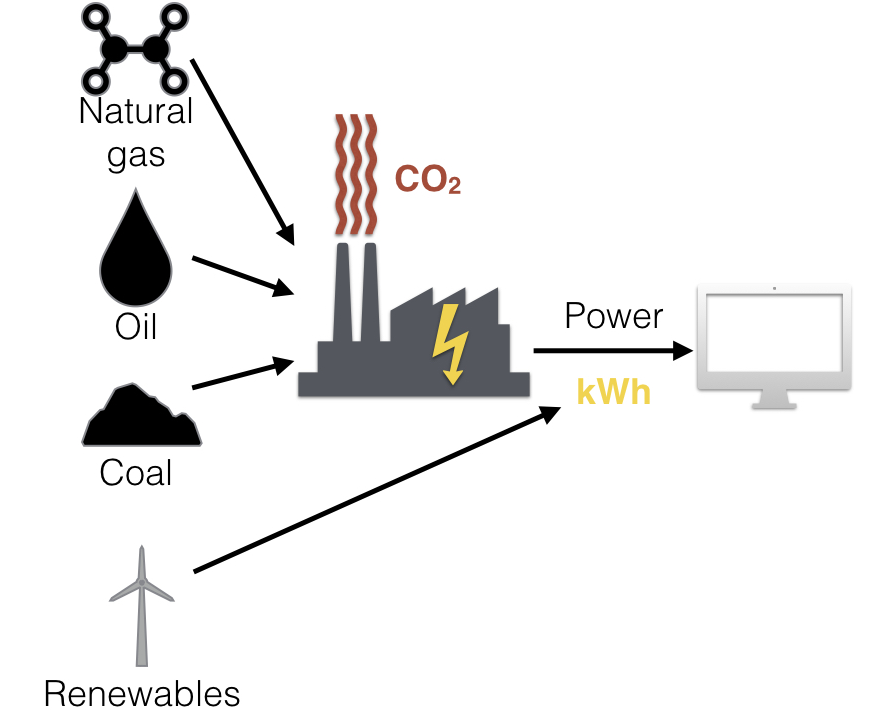}
    \caption{A schematic of energy sources powering local computation. Fossil fuels, particularly coal, oil, and natural gas, are combusted in power plants to generate electricity, emitting carbon dioxide and other greenhouse gases as a byproduct. This electrical power is then transmitted through an electrical grid that powers local computational devices. A diverse suite of renewable or low-carbon energy sources, including hydroelectric power and wind power, also supply the electrical grid. When measuring the environmental footprint of local computation, the types and composition of energy sources for the local electrical grid are therefore critically important.}
    \label{fig:power_transmission}
\end{figure}

The nomenclatures used in the energy and sustainability literatures are often opaque and hinder interdisciplinary comprehension. It's therefore worth explicating the key terminology and building upon them to clarify the energy footprint of computation.
\paragraph{Energy vs. power}
\textbf{Energy} is the amount of work done, or to, an object. Energy is measured in \textit{joules}: one joule is equal to the heat radiated from a current of one ampere flowing through an electrical resistor with a one ohm resistance. As a unit, joules are useful because they can be converted to a related term, \textbf{power}, which is defined as energy per unit time. 
\paragraph{Units of power}The international standard (SI) unit of power is the \textbf{watt}. One watt is defined as one joule per second. At the scale of power consumed by the electrical grid, larger dimensions are important, particularly the \textit{kilowatt} (1,000 watts), the \textit{megawatt} (1,000,000 watts), and the \textit{gigawatt} (1,000,000,000 watts). However, sustainability questions often focus on total energy consumed, rather than the rate at which energy is used, so power can be multiplied by the time interval to yield \textbf{energy consumed}. Frequent units for energy consumption are the \textbf{kilowatt-hour}, or \textbf{kWh}, which is defined as the energy consumed at a rate of one kilowatt for one hour, or 3,600,000 joules (3.6 MJ), or the \textbf{megawatt-hour}, or \textbf{MWh}, which is defined as the energy consumed at the rate of one megawatt for one hour, or 3,600,000,000 joules (3.6 GJ). Power plants frequently express their energy outputs as units of energy consumed: kilowatt-hours (kWh) or megawatt-hours (MWh). Conveniently, environmental footprints of power sources are often expressed as emissions per unit of energy consumed: for example, kilograms of carbon dioxide emitted per megawatt-hour. Therefore, expressing the energy consumption of computational activity in units of energy consumption (e.g., megawatt-hours) can allow the footprint to be calculated directly.
\paragraph{Fuel mixes and the power grid}
The electrical grid is a distribution network of electrical power with power plants as sources. Fossil fuel power plants convert carbon-based fuel in solid, liquid, and/or gas form (e.g., coal, oil\footnote{Technically, the broad category of carbon-based liquid fuels is \textit{petroleum}, which would include many types of products derived from crude oil, including gasoline, diesel fuel, kerosene, and other fuels. However, in both the energy literature and the sustainability literature, \textit{oil is often used interchangeably with petroleum}. Therefore, because oil is likely to be the most precise and most familiar term for computer scientists, we follow this convention here.}, and natural gas, respectively) into energy by combusting these fuel sources to heat water to turn turbines which generate electricity, which is distributed through transmission lines to wide geographic areas. This fuel combustion releases greenhouse gases to the atmosphere, including carbon dioxide. Electrical resistance from transmission lines is high, so long-distance export of electricity between countries is relatively rare, but fuel export between countries (e.g., shipments of coal, oil, and natural gas) is common. Most regions are powered by a mix of power plants that may use different fuels from one another; it's not uncommon for geographic units to have a mixture of different types of fossil fuel power plants (e.g., coal-fired power plants, natural gas power plants) alongside renewable energy plants (e.g., hydropower, solar power). Fuel source choice has a significant impact on the environmental footprint of an individual power plant: for example, among fossil-fuel sources, coal-fired power plants have higher carbon dioxide emissions per unit power produced, followed by oil-burning power plants, and natural gas power plants have the smallest. Therefore, the composition of power sources in the local electrical grid that power an individual computational device will be the most significant factor in determining the carbon emissions footprint of that particular device. Our Python package is designed to make this calculation transparent and translatable for software designers.   

\section{Methodology}

\subsection{Energy Usage Measurement}
\label{sec:method_energy}

To calculate the energy consumption, we use the Running Average Power Limit (RAPL) interface found on Intel processors. The interface allows users to query non-architectural model-specific registers that provide power-related information about the CPU. They are used primarily for limiting power consumption, but the Energy Status register allows for power measurement \cite{Intel_manual}.

The RAPL interface differentiates between several domains based on the number of processors. We use the package domain (or a sum of all of the package domains, on multi-processor machines) as it represents the most accurate overall energy consumption figure for the processor(s).

As outlined by \cite{weaver_rapl}, there are multiple ways to access the RAPL interface data on a Linux machine. We elect to read the files under \url{/sys/class/powercap/intel-rapl} so as to avoid the need for superuser access on the subject machine. These files contain the value of the Energy Status register, which expresses energy used in microjoules, and are updated roughly every millisecond \cite{Intel_manual}. The value in the file increases to the point of overflow and then resets. We take 2 readings with a delay in-between, and then calculate the wattage based on the difference (energy) and the delay (time). To avoid errors due to the reset of the file, we discard negative values.

To the measurement obtained via the RAPL interface we additionally add the power usage of the GPU for machines with an Nvidia GPU which supports the NVIDIA System Management Interface program. Similar to RAPL, NVIDIA-smi is a utility that allows the user to query the current power usage of the GPU. Since the power usage is expressed in watts, we simply add it to the CPU measurement.

Another consideration is power supply efficiency which is defined as output power divided by input power. It demonstrates the amount of energy lost to heat when powering the machine. The issue with incorporating this value into the package is twofold: there is no way to detect the efficiency from the command line and there is variance in the efficiency among different power supplies. However, there is a voluntary certification program called 80-Plus which indicates the efficiency at different loads. As indicated by the name, it certifies that the power supply has at least 80\% efficiency at loads of 20\%, 50\%, and 100\%. Thus, we allow users to specify the efficiency, if known, and otherwise default to 80\%.

\subsection{Calculating CO2 Emissions}
\label{sec:method_emissions}

\paragraph{Location}
In order to accurately calculate the CO$_2$ emissions associated with the computational power used, we firstly determine the geographic location of the user via their IP address with the help of the GeoJS API \cite{geojs_api}. If the location cannot be determined, we use the average energy mix and carbon emissions of the world as the default. This average world energy mix is 28.7\% coal, 22.9\% oil, 33.9\% natural gas, and 14.4\% low-carbon fuels with a resulting world national average CO$_2$ electricity emissions  of 1600.6 lbs CO$_2$ per MWh (= 726 kg CO$_2$ per MWh) \cite{USEIAdata}.  Package options are also provided to allow the user to instead specify using a European or US average if the location is unknown.

\paragraph{Data}
Our United States energy mix and emissions data was obtained from the U.S. Environmental Protection Agency eGRID data for the year 2016 \cite{eGRID_summary_table_file}. We used the State Resource Mix section for displaying the energy mix, and the State Output Emission Rates section for calculating emissions in the United States.\footnote{The eGRID data breaks down fossil fuel energy sources into numerous categories. This analysis focused on three categories: coal, "total petroleum" (oil), and natural gas. We did not use the ``otherFossil" data shown in eGRID because the values were predominantly 0 (and in cases in which the value was nonzero, it was below 1\%). These values often refer to small-scale electricity generation through the combustion of wood or other biomass, such as peat.}

We obtained international energy mix data from the U.S. Energy Information Administration data for the year 2016 \cite{USEIAdata}. Specifically, we looked at the energy consumption of countries worldwide, broken down by energy source. We removed from consideration former countries for which no data was available (Former Czechoslovakia, Former Serbia and Montenegro, Former U.S.S.R., Former Yugoslavia, Hawaiian Trade Zone, East Germany and West Germany), and approximated to 0 for data points with negligibly small values.

As of July 2019, the most recent eGRID data was from the year 2016. We elected to use 2016 U.S. E.I.A. data for consistency between the data sources.

\paragraph{Conversion Rates}
In order to convert the energy used (determined via the RAPL interface; see Appendix \ref{sec:method_energy}) into CO$_2$ emissions, we were forced to take two different approaches based on the difference in data sources that are available. For the United States, the eGRID data contains individual states' output emission rates measured in pounds of carbon dioxide emitted per MWh of electricity consumed. Thus, to calculate kilograms of carbon dioxide emitted per kilowatt-hour, we simply perform the conversion from MWh to kWh and from pounds to kilograms. International carbon emissions footprints are more complicated: because the most reliable international data is simply the energy mix found in that location (e.g., what proportion of electricity is generated from coal, oil, and natural gas), we had to ensure that the conversion from each specific energy source to amount of CO$_2$ emitted was as accurate as possible. An issue we encountered when performing initial conversions was that our conversion formulas to CO$_2$ were used in broader contexts of energy consumption, and thus less accurate. To that end, we elected to reverse-engineer the formulas for the carbon footprint of each individual \textit{fuel source} from the eGRID data (using both state emission output and state energy mix), which have been found to be reliable, and then applying that formula to the energy mix of each country. This approach not only provides us additional consistency between the datasets, but it also ensures that the conversions pertain to the specific domain of electricity consumption. 

\paragraph{Calculating Carbon Footprints of Fuel Sources}
The state-level energy production data for the United States contains two key values: the energy production (in MWh) from each of the three carbon fuel sources (coal, oil, and natural gas); and the thousands of metric tons of CO$_2$ generated from each particular source (coal, oil, and natural gas). Our goal was to determine how many kilograms of CO$_2$ were generated per MWh for each of the three fuel sources. To reach this value, we converted the value of carbon emissions presented (in thousands of metric tons) into kilograms (1 metric ton = 1,000 kilograms) of CO$_2$ by multiplying the state-level emissions value by 1,000, yielding kg of CO$_2$ for that fuel source. Dividing the kilograms of CO$_2$ by the energy production of the particular source involved (e.g., coal, oil, and natural gas) yielded a carbon emission intensity for coal, oil, and natural gas, in units of kg CO$_2$ emitted per MWh of power generated. This value was then used to measure the carbon footprint of international power grids, because values of energy production (e.g., MWh from coal, oil, and natural gas) are reliable and easily found. 
The conversion calculations are therefore:
\begin{equation}
metric\:tons\:CO_2\;x\;1,000 = kg\:CO_2
\end{equation}
\begin{equation}
Emissions = \frac{kg\:CO_2}{MWh}
\end{equation}

Energy efficiencies of power plants vary based on local equipment, so to establish an average carbon intensity for coal, oil, and natural gas, we compared values for three US states (West Virginia, Missouri, and Wyoming) to yield an average for each of the three fuel types. These three states were chosen because they had a diverse set of fossil fuel power plants and an energy infrastructure with a mix of younger and older power plants. Carbon emissions per unit power generated were highest for coal  (934, 975, 1085 kg CO$_2$/MWh), followed by oil (735, 922, 798 kg CO$_2$/MWh), and then natural gas (700, 528, 1009 kg CO$_2$/MWh), although variance was highest in natural gas, followed by oil, and smallest for coal. We used the mean values for these three states (coal: 996 kg CO$_2$/MWh; oil: 817 kg CO$_2$/MWh; natural gas: 744 kg CO$_2$/MWh) to calculate the carbon emissions from international data. These values are fixed for all countries to keep values consistent between the United States and other countries.

\section{Package Validation and Usage}

\subsection{Validation}
\label{sec:validation}

\begin{figure*}
    \centering
    \includegraphics[height=1.5in]{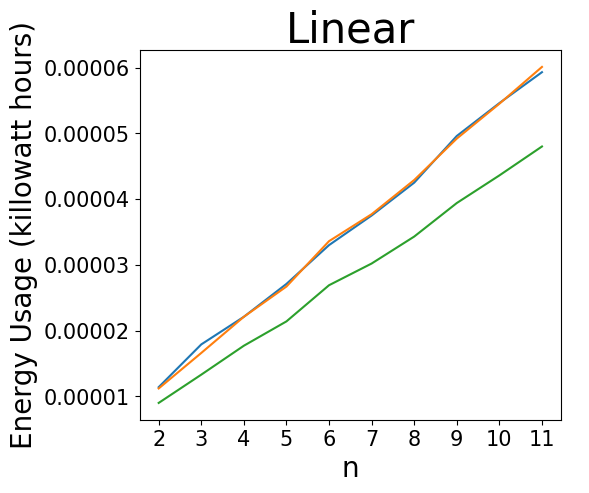}
    \includegraphics[height=1.5in]{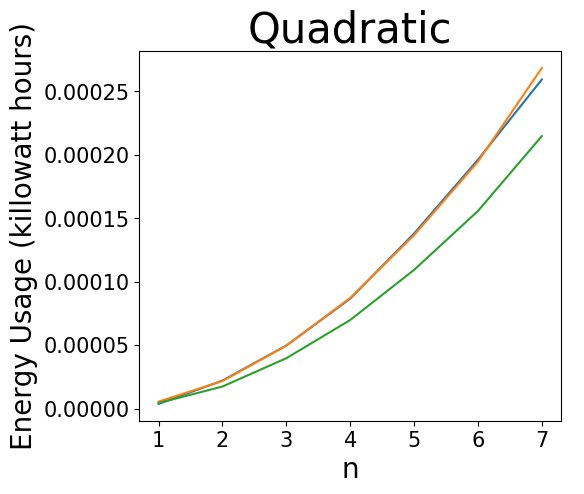}
    \includegraphics[height=1.5in]{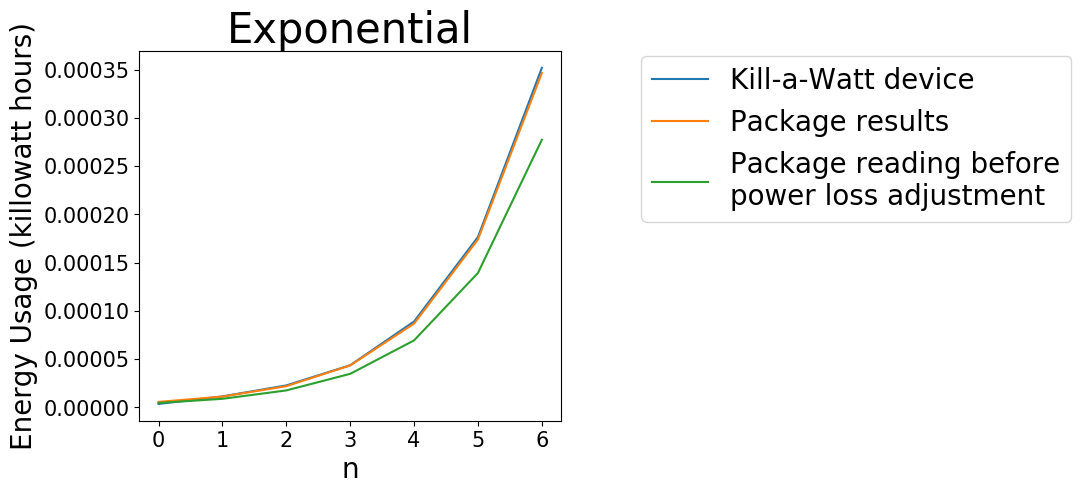}
    \caption{Validation experiments showing that, across functions with varying complexities, the energy measurement device readings match the power loss adjusted readings from the \energyusage\ package.}
    \label{fig:measurement_validation}
\end{figure*}

\begin{figure*}[htbp]
\begin{multicols}{2}
\begin{verbatim}
    def linear(n):
        for i in range(n):
            for j in range(50000000):
                num = 1+1
    def quadratic(n):
        for i in range(n):
            for j in range(n):
                linear(1)
    def exp(n):
        for i in range(2**n):
            linear(1)
\end{verbatim}
\columnbreak
\begin{verbatim}
import energyusage

# user function being evaluated is exp(n)
# with n=10 in this example
# optional argument pdf=True writes the
# Energy Usage Report out to file

energyusage.evaluate(exp, 10)

# returns:
# kilowatt hours used, function return values
# prints out Energy Usage Report
\end{verbatim}
\end{multicols}
\caption{Left: Functions used for energy measurement validation.  Right: Basic \energyusage\ package usage.}
\label{fig:code}
\end{figure*}

In order to validate the above methodology for programmatic energy usage measurement, we used a Kill-a-Watt energy usage monitor. The computer running the energy usage tests was plugged into the monitor. Our goal was to consider the energy usage as calculated programmatically in comparison to the energy usage calculated directly by observing the power draw using a physical monitor.

In order to validate the energy usage measurement over a variety of complexity classes, we then developed a program with a controlled number of additions (a function that simply adds 1 a specified number of times). All experiments were run on an Intel Core i5-7500 CPU (3.40GHz).  On this computer, $50,000,000$ additions took roughly one second. The \emph{linear} function measured was $n \cdot 50,000,000$ additions, the \emph{quadratic} function did $n^2 \cdot 50,000,000$ additions, and the \emph{exponential} function did $2^n \cdot 50,000,000$ additions (functions shown in Figure \ref{fig:code}).

For each value of $n$, the function was run and energy usage was recorded using the \energyusage\ package as well as the Kill-a-Watt device. Kill-a-Watt recordings were taken manually by per-second observation (the device has a screen display). Since the readings on the device were found by ad-hoc experimentation to suffer from a one or two second delay, the first one or two such manual readings were discarded (determined based on an observed spike in energy usage). The remaining Kill-a-Watt readings during the run of the function were averaged and converted to kWh using the time for each process as recorded by \energyusage.

The resulting energy usage recordings per varying values of $n$ and the different measurement methods are shown in Figure \ref{fig:measurement_validation}. The package-based measurements are shown both before and after the power loss adjustment, which was set to $0.8$ efficiency.  Based on these experiments, it is clear both that the $0.8$ value was accurate for the computer used and that the \energyusage\ package measurement successfully matches that of the Kill-a-Watt monitoring device while being entirely programmatic.

\subsection{Package Usage}

Our main goal in the development of the package was to make reporting the energy usage of a function as easy for the user as possible. Thus, the package handles all of the assumptions and calculations detailed above for the user.  The main function, \texttt{evaluate}, takes another function and that function's parameters as input. It then runs the given function and evaluates its energy usage, returning both the function's return value(s) and, optionally, the amount of kilowatt-hours used. It also has an optional argument that allows the report to be directly written to a PDF; by default, the report is written out to the command line. Another optional argument allows the user to specify the power supply loss, if known, for their computer. The default power supply loss value is $0.8$ since 80Plus energy certification has now become commonplace. See Figure \ref{fig:code} for example usage of the package.

\section{Energy Usage Reports: Detailed Description}
\label{sec:report_desc}

In this section we detail and motivate each of the sections included in the produced \report s.  An example \report\ for the simple \texttt{exp} function shown in Figure \ref{fig:code} is given as Figure \ref{fig:report}.  Many other aspects of algorithmic accountability should also be reported; we refer the reader to other work for suggested reporting mechanisms on non-environmental impacts of algorithms (\cite{gebru2018datasheets,mitchell2019model, selbst2017disparate, holland2018dataset, diakopoulos2016principles, reisman2018algorithmic, yang2018nutritional}).

\subsection{Energy Usage Readings}
The average baseline wattage, total wattage, and process wattage (equal to average total wattage minus average baseline wattage) are displayed at the top of the report, along with the duration of the process. Each of these readings are important for completing the environmental footprint calculation. The average baseline wattage displays how much power is being used by the local computational device when it is at rest; the average total wattage indicates the amount of energy used; and the average process wattage (defined as $\mbox{\texttt{total wattage}} - \mbox{\texttt{baseline wattage}}$) is how much energy was used by the function investigated in the report. The process duration is displayed because, as described in Section \ref{sec:background}, the time interval of the calculation must be known in order to determine the total energy consumed. Using the process wattage and process duration, the total energy consumed (in kWh) can be calculated.  Thus, these basic readings are what would be necessary to reproduce the report in the case of future changes to the energy mix data or assumed carbon equivalencies.

\subsection{Energy Mix Data}
As described in Section \ref{sec:background}, the types of energy used to power the local grid and their relative amounts are critical for determining the carbon footprint of local calculations. In this analysis, the fuels with the highest carbon dioxide intensity (coal, oil, natural gas) are represented on the pie chart, with the amount of low-carbon fuels (for example, hydroelectricity) also represented.  Making the energy mix for a specific named location explicit and visible as part of the report allows for broader understanding of the importance of the local grid to \co\ emissions.

\subsection{Summary Results}
The center of the \report\ shows a highlighted box with the total kilowatt hours used and the calculated effective \co\ emissions for the program. These two values are the key reporting-out numbers that facilitate an understanding of the environmental impact of local computation. Carbon emissions (expressed here in units of kg \co) are the unit of environmental impact and can be compared with other environmental variables (e.g., emissions from automobiles or agriculture). Reporting energy consumed (expressed here in units of kWh) facilitates comparison with other energy-intensive activities. 

\subsection{Assumptions}
Under ``Assumed Carbon Equivalencies" we show the particular scalars used to quantify \co\ emissions per megawatt-hour by each source of electricity, as described further in Section \ref{sec:method_emissions}. For the purposes of this exercise, we model low-carbon fuels as having 0 emissions. Although some types of low-carbon power generation facilities have been found to produce greenhouse gas emissions as a consequence of their architecture or design \cite{Deemer_GHG_dams}, carbon emissions from fossil fuel power plants dominate the greenhouse gas footprint of the energy sector.

Further extensions of the work presented here could account for carbon emissions caused by the construction of the electricity grid (accounting for, for example, \co\ emissions from the construction of coal-fired power plants along with carbon dioxide emitted during the construction of solar panels) but that is beyond the scope of this paper and package, which focuses on the emissions caused by the consumption of electricity by local computation in real time.  We hope that the statement of these assumptions as part of the \report\ helps provide full reproducibility and transparency in light of the possibility of future improvements to these assumptions.

\subsection{CO2 Emissions Equivalents}
We contextualize the emissions from this analysis by comparing the carbon dioxide values with values familiar to people from their everyday life: automobile miles driven, minutes of television watched on a 32" LCD screen, and as the percent of \co\ emitted by a typical household in the United States during an average day. Further analysis could contextualize the calculated footprint of computation in different ways that are appropriate to the local audience (for example, kilometers traveled). Other metrics that could be used to contextualize the emissions equivalents can be found elsewhere \cite{carbon_footprint_book}.  We believe that such contextualization is critical to encouraging a human understanding of the impact of energy usage and emissions.

\subsection{Emissions Comparisons Worldwide}
Since the fuel sources used to power the local electricity grid vary between geographic locations, it is instructive to compare what the carbon emissions would be for the same calculation if it were powered by different energy grids. For example, the footprint of any calculation would be greatly diminished if the local electricity grid is dominated by low-carbon fuels, or greatly amplified if the local electricity grid is dominated by coal-fired power plants. Therefore, this section of the \report\ illustrates the carbon dioxide emissions of the same calculation powered by different electrical grids from the United States, Europe, and worldwide (minus the United States and Europe).

Each of the three panels shows the \co\ emissions of the function powered by the local grid when compared with the lowest, median, and highest emissions grid for that region. The leftmost plot compares the carbon dioxide emissions of the function with the emissions that would result if it were powered by the lowest-emitting state (Vermont), the median-emitting state (Mississippi), and the highest-emitting state (Wyoming). The central plot follows the same pattern but with the lowest-emitting (Iceland), median-emitting (Ukraine), and highest-emitting (Kosovo) European countries.\footnote{We use the most broad definition of Europe which also includes countries such as Azerbaijan, Georgia and Kazakhstan.} The rightmost plot includes the global (excluding the United States and Europe) context, with the lowest-emitting country (Bhutan), the median-emitting country (South Korea), and the highest-emitting country (Mongolia). The variation found across the globe is dominated by countries whose electricity grid contains substantial low-carbon fuels (e.g., hydropower) versus other countries whose electricity grid is powered by carbon-intensive fossil fuels (e.g., coal, oil). Future analysis will represent these plots graphically through maps.

The \report\ is, thus, designed to encourage algorithmic accountability by allowing for full reproducibility and clear statement of assumptions, as well as by raising the visibility of key emissions equivalents and global comparisons.  We anticipate that in-depth analysis of specific algorithms spurred by the results shown in the report will focus on comparisons using the highlighted summary results. We give one such example analysis in the next section.

\subsection{Experimental Setup}
\label{sec:experimental_setup}
In order to assess this potential trade-off, we considered four standard models (support vector machines, $k$-nearest neighbors, Ada Boost with a decision tree base estimator, and a multi-layer perceptron) on two different datasets. The first dataset is the Adult Income dataset \cite{Dua:2019} containing $48,842$ instances of individuals' census data used to predict whether they make more or less than $50,000$ per year.  The second dataset is the ProPublica dataset \cite{angwin2016machine} containing data about $7,215$ individuals arrested and used to predict whether they will be rearrested within two years. Both datasets are preprocessed by removing sensitive variables (race and sex) and one-hot encoding categorical variables. The ProPublica dataset is additionally preprocessed as described in the original study \cite{angwin2016machine}. A two-thirds to one-third training versus testing split is created.

For each of the four types of trained models, we adjusted a parameter across a range of values to allow the classifiers to be tuned.  (Other than these parameters, the \texttt{sklearn} package defaults were used for all parameters.)  In the case of some classifiers, e.g., Ada Boost, this parameter (the number of estimators) is one that is likely to increase the quality of the model at the cost of time and energy usage. In the case of the $k$-nearest neighbors classifier, we adjusted the value $k$ of the number of nearest neighbors to consider - a parameter that is far less likely to correlate with either energy usage or accuracy.

\end{document}